\documentclass[dvipsnames]{article}

\usepackage[utf8]{inputenc}
\usepackage[T1]{fontenc}
\usepackage[USenglish]{babel}
\usepackage{csquotes}

\usepackage{amsmath}
\usepackage{amsfonts}
\usepackage{color}
\usepackage{placeins}

\newcommand{\Tau}{\mathrm{T}}
\newcommand{\truerisk}{\rho}

\usepackage{hyperref}
\usepackage{cleveref}
\usepackage{graphicx}
\usepackage{layouts}

\usepackage[a4paper, total={5.8in, 8.5in}, footskip=40pt, footnotesep=20pt]{geometry}
\usepackage{authblk}

\makeatletter
\renewcommand\AB@authnote[1]{\rlap{\textsuperscript{\normalfont#1}}}

\makeatother

\usepackage{fancyhdr}
\usepackage[ttscale=.85]{libertine}
\usepackage{libertinust1math}
\usepackage[activate={true,nocompatibility}, final, tracking=true, kerning=true, spacing=true, factor=1100, stretch=10, shrink=10]{microtype}
\microtypecontext{spacing=nonfrench}

\SetExtraSpacing{ encoding = {TS1} } { }

\usepackage{fnpct} 

\usepackage[
backend=bibtex,
style=numeric-comp,
maxcitenames=2,
natbib
]
{biblatex}

\emergencystretch=\maxdimen
\usepackage{xcolor}
\hypersetup{
    colorlinks,
    linkcolor={red!50!black},
    citecolor={blue!50!black},
    urlcolor={blue!80!black}
}

\begin{document}

\title{On (assessing) the fairness of risk score models}

\author[1, 2, *]{Eike Petersen}
\author[2, 3, 4]{Melanie Ganz}
\author[2, 5]{Sune Hannibal Holm}
\author[1, 2]{Aasa Feragen}
\affil[1]{Technical University of Denmark, DTU Compute}
\affil[2]{Pioneer Centre for AI, Denmark}
\affil[3]{University of Copenhagen, Department of Computer Science, Denmark}
\affil[4]{Neurobiology Research Unit, Rigshospitalet, Copenhagen, Denmark}
\affil[5]{University of Copenhagen, Department of Food and Resource Economics, Denmark}
\affil[*]{Corresponding author. Contact: \href{mailto:ewipe@dtu.dk}{ewipe@dtu.dk}.}
\date{}

\maketitle

\begin{abstract}
Recent work on algorithmic fairness has largely focused on the fairness of discrete decisions, or classifications. 
While such decisions are often based on risk score models, the fairness of the risk models themselves has received considerably less attention. 
Risk models are of interest for a number of reasons, including the fact that they communicate uncertainty about the potential outcomes to users, thus representing a way to enable meaningful human oversight. 
Here, we address fairness desiderata for risk score models. 
We identify the provision of similar epistemic value to different groups as a key desideratum for risk score fairness. 
Further, we address how to assess the fairness of risk score models quantitatively, including a discussion of metric choices and meaningful statistical comparisons between groups. 
In this context, we also introduce a novel calibration error metric that is less sample size-biased than previously proposed metrics, enabling meaningful comparisons between groups of different sizes. 
We illustrate our methodology -- which is widely applicable in many other settings -- in two case studies, one in recidivism risk prediction, and one in risk of major depressive disorder (MDD) prediction.
\end{abstract}

\section{Introduction}
To date, much of the algorithmic fairness literature has focused on the fairness of classification systems which are used, for example, to decide whether a person should be granted a loan or be released from prison on bail.
Even in cases where such classification decisions are based on risk score models -- such as in the highly influential COMPAS case~\cite{Chouldechova2017, Dressel2018, Bao2022} -- their fairness is typically considered a function of the decisions, or \emph{classifications}, made by the system.
Of course, any risk score model can be turned into a classifier by selecting a probability threshold (in binary classification) or predicting the most likely outcome (in multi-class classification).
Nevertheless, we argue here that it is worthwhile to distinguish between these two settings and consider the fairness of risk models independent of their downstream use, be it as the basis for a classifier or otherwise. 
We discuss notions of fairness for risk scores as well as their relationship to classical, classification-level notions of fairness, and we develop robust tools to empirically quantify risk score fairness.
We illustrate our methodology in two case studies, one situated in the criminal justice system and one in healthcare.

\subsubsection*{Why distinguish between fair models and fair decisions?}
In the statistical literature, it is generally considered desirable to distinguish between inference (e.g., identifying a risk score model) and subsequent decision-making (e.g., deriving a classification from a risk score model): while the former represents a purely statistical task, the latter depends on subjective choices about the relative desirability of different outcomes~\cite{Jaynes2003, Rambachan2020}.
Similar arguments have been made in the context of fairness analyses: learning a fair prediction model and making fair decisions based on a model are two separate tasks that should be considered separately~\cite{Hedden2021}.
At the same time, risk score models also play an important role outside the realm of automated decision-making, as a tool for human-computer interaction.
Risk scores communicate \emph{uncertainty} about the potential outcomes to users, thus enabling meaningful human oversight and decision-making as required for high-risk AI applications, e.g., by the EU's proposed AI Act~\cite{EUCommission2021}.
For this reason, risk scores have a long and successful history in fields such as medicine~\cite{Kattan2016, Adeyemo2021, Zhao2020a, Noble2011, Patel2009, Kutz2016}, lending further urgency to an analysis of their fairness outside the realm of automated classification systems.

\subsubsection*{Contributions.}
Our contributions range from the conceptual to the technical level, motivated by a single question:
\begin{center}
\emph{How can we define and reliably quantify the fairness of risk score models?}
\end{center}
In particular, our contributions include the following:
\begin{enumerate}
\item We provide a principled argument for equality of \emph{epistemic value} provided to different groups as a key desideratum for the fairness of risk models (\cref{sec:desiderata}). 
By epistemic value, we here refer to the utility of the model in predicting the true risks of subjects. 
From this, we derive the relevance of previously proposed metrics, such as calibration, as well as new ones, for the fairness of risk scores.
In particular, we argue that calibration and discriminative ability, that is, how well a model is able to separate positive from negative examples, are two complimentary key properties of interest.
The latter can be measured, for example, by the area under the receiver-operating characteristic curve (AUROC) or the area under the precision-recall-gain curve (AUPRG).
\item We examine whether risk score models can be used to obtain \emph{fair rankings} in, e.g., scarce resource allocation scenarios such as medical triage (\cref{sec:ranking}). We show how even otherwise fair risk score models can lead to unfair rankings, and we show how ranking fairness relates naturally to demographic parity, or independence, in classification models. Following this analysis, we discuss consequences and make recommendations for the use of risk score models in resource allocation and prioritization.
\item We address the \emph{quantification of potential fairness violations in practice} (\cref{sec:metrics}).
In this context, we discuss the importance of metric uncertainty quantification in order to avoid premature conclusions regarding the fairness or unfairness of a model.
In addition, we introduce a \emph{novel calibration error metric} that is more reliable and less sample size-biased than existing, commonly used metrics.
The latter property is of particular relevance in the fairness context, where groups of different sizes are frequently compared.
\item We illustrate our methodology -- which is widely applicable -- using two \emph{case studies}, one in juvenile recidivism risk prediction, and one in risk of major depressive disorder (MDD) prediction (\cref{sec:case-studies}).
While the first case study uses a previously published and analyzed dataset~\cite{Tolan2019}, the second case study, based on national registry data of more than 400,000 subjects, is completely new.
\end{enumerate}

\subsubsection*{Notation and problem setting}
We consider the fairness of a risk score~$R\in [0, 1]$ computed by a model $f(X)$ from input features~$X$.
We make no assumptions about~$f$ -- it could be, for example, a simple linear model, a neural network, or a random forest model.
A risk score is designed to approximate the \emph{true risk}~$\truerisk=P(Y=1 \mid X)$ of an outcome $Y\in\{0, 1\}$ being positive ($Y=1$), given the input features~$X$.
We refer to this as the \emph{fundamental aim of risk modeling}.
In some places, we will consider binary predictions $\hat{Y}\in\{0, 1\}$ derived from risk scores~$R$ by means of a decision threshold~$\tau\in[0, 1]$.
We pursue a notion of \emph{group fairness}, comparing model properties between subject groups~$G$ derived from protected attributes~$X_p$.
The protected attributes may or may not be partially or fully contained in the model inputs $X$.
We consider categorical groups only; continuous-valued sensitive attributes (such as age) are discretized into a set of bins.
Groups of interest may be defined by one or more sensitive attributes, and groups may overlap.

\section{Background and related work}
\label{sec:background}
While the fairness of risk models has been widely discussed in the literature, this has been predominantly done in the context of risk scores used for automated decision-making, i.e., discrete classifications derived from risk scores based on a fixed threshold~\cite{Chouldechova2017, Hardt2016a, Fogliato2020, Kallus2018, Kearns2018, Wang2021a, Pfohl2022a}.
As a consequence, many of the proposed standard fairness metrics such as equalized odds, equal opportunity, parity of predictive values, or net benefit, are a function of classification \emph{decisions}, and not of the risk score itself~\cite{Hardt2016a, Pfohl2022a}.
This can be problematic for a number of reasons.
First, decision thresholds might not be known in advance or might vary over time, thus inducing a need for an assessment that considers a whole range of possible thresholds.
Second, any such classification-based fairness metric is based on implicit assumptions about the relative costs and benefits of different true or false decisions, which might not be known in advance.
Third, separating between the aim of learning a \emph{fair model} and making \emph{just decisions} enables a more precise investigation into the sources of potential unfairness~\cite{Hedden2021}.
And, finally, there might simply be no formal decision-making process to analyze, for example in cases in which risk scores are used to inform human decision-making.

There is one notable exception to this focus on the fairness of classifications in the context of risk scores:
\emph{calibration by groups}, which is one of the standard fairness metrics employed in this space~\cite{Kleinberg2017, Chouldechova2017, barocas-hardt-narayanan}, is indeed a risk score-level metric that is fully independent of any subsequent decision-making. 
Calibration by groups is defined as~\cite{barocas-hardt-narayanan}
\begin{equation}
    P(Y=1 \mid R=r, G=g) = r \quad \forall \, r, g.
    \label{eq:calibration}
\end{equation}
In words, the model's risk predictions should correspond to the actual risk in the population --  a property of central importance to the project of risk modeling.
Calibration as a desirable property of probabilistic models has a long history in diverse fields~\cite{Pettigrew2019, Calster2019, Hedden2021}.
Many different metrics have been proposed for assessing calibration performance, including the expected calibration error (ECE)~\cite{Naeini2015}, adaptive calibration error (ACE)~\cite{Nixon2019}, and various metrics based on proper scoring rule decompositions~\cite{Dimitriadis2021, Murphy1973, Broecker2009, DeGroot1983}.
Unfortunately, it has been noted that standard calibration error estimates suffer from severe sample size bias~\cite{Broecker2011, Kumar2019, Roelofs2022, Gruber2022}, thus necessitating the development of less biased alternative metrics.
Towards this end, \citet{Kumar2019} proposed a binned version of the bias-corrected Brier score decomposition described by \citet{Ferro2012}, corresponding to a debiased estimate of the root mean squared calibration error.
They used equal-width binning and a fixed number~$N_B$ of 15 bins.
Later, \citet{Roelofs2022} found that bias could be further reduced by using equal-mass binning. 
In addition, \citet{Roelofs2022} describe a variant of ACE in which they choose the number~$N_B$ of bins adaptively as the highest bin number that still yields a monotonically increasing calibration curve.
They find this method to perform similarly well compared to an equal-mass binning version of the debiased calibration RMSE described by \citet{Kumar2019}.
Interestingly, however, \citet{Roelofs2022} do not apply their proposed bin count search method to the metric described by \citet{Kumar2019}.
Moreover, to the authors' knowledge, the implications of sample size-biased calibration error metrics for fairness analyses have not yet been discussed.
In the fairness context, such biases are especially problematic, as groups of different sizes are routinely compared.

In the context of predictive modeling, calibration is often considered in conjunction with a second desirable property, \emph{discriminative ability}~\cite{Steyerberg2010}.\footnote{This property has been the subject of many definitions across fields, including \emph{refinement}~\cite{DeGroot1981, Kull2015}, \emph{sharpness}~\cite{Gneiting2007a}, \emph{separation}~\cite{Flach2008}, and \emph{resolution}~\cite{Murphy1973}, all of which describe very closely related and, in many cases, identical concepts.}
Maybe most prominently, discriminative ability can be measured by the area under the receiver-operating characteristic curve (AUROC)~\cite{Bamber1975, Flach2011, Flach2016}.
The AUROC metric is of particular relevance in our setting as it does not rely on binary classification decisions, quantifying the accuracy of the score \emph{ranking} instead.
This can also be understood from its equivalence to the \emph{concordance statistic}~\cite{Hanley1982, Steyerberg2010}, i.e., the likelihood of a randomly selected positive sample being ranked above a randomly selected negative sample.\footnote{Another related result is the equivalence between AUROC and the Mann-Whitney U statistic~\cite{Mason2002}.}
It has been noted that in the case of strong class imbalance, AUROC can be misleading~\cite{Saito2015}, and in such cases, the area under the precision-recall curve, equivalent to the average precision (AP), has been proposed as a potential alternative metric~\cite{Saito2015, MaierHein2022}.
As has been pointed out by \citet{Flach2015}, this metric requires proper normalization in order to enable meaningful comparisons between groups with different base rates, resulting in the use of precision-recall-\emph{gain} curves and the area below them (AUPRG).

Finally, \emph{fair ranking} has also been the subject of many publications, including rankings derived from risk score models; for an overview refer to the recent survey of \citet{Zehlike2022}.
Much of this research is in recommendation systems, however, in which relative position in a ranking matters due to declining attention being given to lower ranks~\cite{Zehlike2022a}.
This differs importantly from the \emph{set selection} setting we consider here, in which relative position is irrelevant as long as a subject is selected for inclusion.
Set selection fairness is intimately tied to ROC curve analysis, as both depend on the distributions of risk predictions in different groups.
\citet{Vogel2021} describe a general ROC-based framework for fair ranking, which subsumes different variations of AUROC fairness, as well as \emph{pointwise} ROC fairness, corresponding to the requirement that, e.g., at every threshold in a given range, true and/or false positive rates should be equal across groups.
\citet{Zehlike2022b} propose a fair top-k ranking algorithm for overlapping protected groups, which ensures a certain minimum representation of each group in the selected set while maximizing overall ranking utility.
The relationship between such approaches and (fair) risk score-based set selection has not analyzed, however.

\section{Fairness desiderata for risk score models}
\label{sec:desiderata}
What should we ask of a ``fair'' risk score model?
Recall that the fundamental aim of risk modeling is to accurately predict the true underlying risks.
If we consider the fairness of a risk model detached from any subsequent decision-making, then we can only characterize its fairness in terms of how well it performs according to this aim in different groups.
We thus argue here that the fairness of a risk model is fundamentally determined by the \emph{epistemic value} it provides to the different groups, and we consider a risk score model \emph{fair} if it provides similar epistemic value to all groups of interest.
This differs fundamentally from the classification case, in which fairness is typically characterized in terms of the relative costs or benefits provided to different groups as a result of \emph{decisions} made based on the model, such as in equalized odds~\cite{Hardt2016a}, equality of opportunity~\cite{Hardt2016a}, or the expected net benefit of an intervention for different groups~\cite{Pfohl2022a}.

Quantifying the epistemic value of a given model or belief has been the subject of intense research and debate for many decades across disciplines.
Interestingly, many of these diverse strands of research converge on \emph{calibration} as a key desirable property of a model from an epistemic point of view~\cite{Pettigrew2019, Calster2019, Hedden2021}.
In practice, calibration will never be satisfied exactly, and much research has been devoted to assessing the degree to which a model is miscalibrated; refer to \cref{sec:background} and \cref{ssec:calibration} for more details on this issue.
For now, we simply state that we consider \emph{similar degrees of (mis-)calibration across groups} our first fairness desideratum for risk score models.

However, a calibrated model need not be epistemically valuable.
A model that always returns a group's base rate $P(Y=1 \mid G)$, regardless of any input features~$X$, is perfectly calibrated but provides little epistemic value.
Thus, it has long been known that a secondary requirement for a model to provide epistemic value is for it to successfully separate positive from negative examples.
We will employ the term \emph{discriminative ability}\footnote{In order to avoid any confusion caused by an unfortunate clash of terminology in the fairness context, we emphasize that \emph{discriminative ability} is a desirable model property and not related to \emph{unfair discrimination}.} to characterize this property, and we consider \emph{similar degrees of discriminative ability across groups} to be our second fairness desideratum for risk score models.
We will return to the issue of specific metrics in \cref{sec:metrics} but would like to make two general remarks here.

\subsubsection*{Calibration and discriminative ability are fully orthogonal properties.}
A model can be perfectly calibrated yet have no discriminative ability, as mentioned above.
The inverse is also true:
Consider a model that always returns~$0.5-\varepsilon$ for negative examples and $0.5+\varepsilon$ for positive examples -- such a model would be maximally miscalibrated while being perfectly discriminative.
We will want our chosen metrics to reflect this orthogonality, i.e., a calibration metric should be independent of discriminative ability, and vice versa.
As a negative example, it is well-known that proper scoring rules such as the Brier score and the log loss measure a combination of calibration and discriminative ability~\cite{Kull2015}, thus being unsuitable for use as a pure measure of either property.
As a positive example, the AUROC metric conforms with this requirement, since it is fully invariant to (monotonic) calibration: as long as a calibration procedure does not change the \emph{ranking} of the returned risk scores, AUROC will be the same before and after calibration.

\subsubsection*{Compatibility of equal calibration and discriminative ability.}
\begin{figure}[t]
    \includegraphics[width=\textwidth]{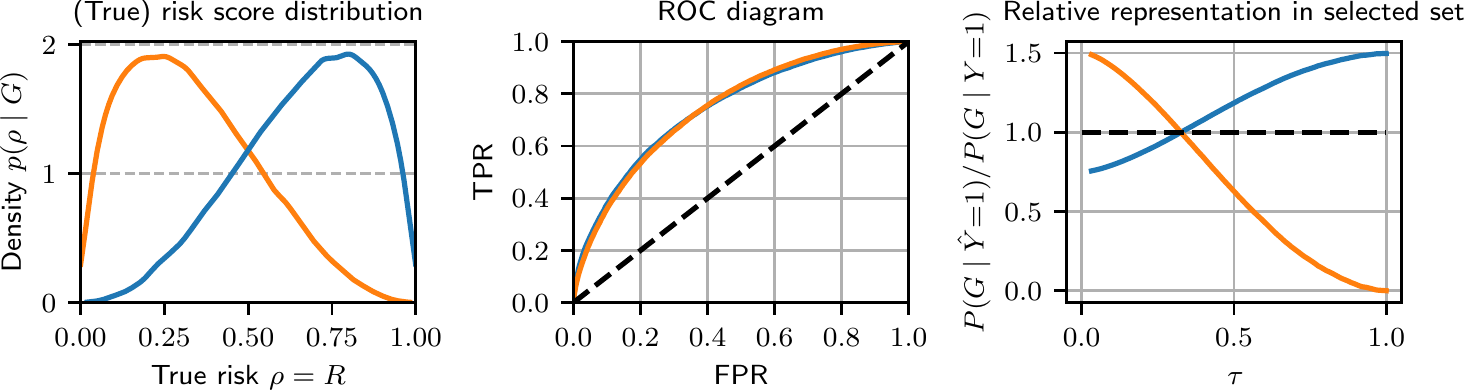}
    \caption{A simple example illustrating how a model can be well-calibrated across groups and AUROC-fair (while violating separation), and how unfair ranking can arise even from such well-calibrated and AUROC-fair models.
    The left panel shows the distributions of true risks~$\tau=p(Y \mid X,G)$ in two groups (blue and orange, assumed equally large).
    Their respective base rates are $p(Y=1 \mid G=\text{blue})=2/3$ and $p(Y=1 \mid G=\text{orange})=1/3$.
    We assume that the true risks are perfectly captured by the model, i.e., $R=\tau$ for each subject.
    Thus, the model is perfectly well-calibrated in both groups.
    (Note that $R=\tau$ is not a necessary requirement for a model to be simultaneously perfectly well-calibrated and AUROC-fair.)
    Both groups have identical ROC curves and AUROCs.
    Nevertheless, the model violates separation ($R \perp G \mid Y$), and enforcing the model to achieve separation would render calibration by groups and the prediction of the true risks impossible~\cite{barocas-hardt-narayanan}.
    Thus, we argue (\cref{sec:desiderata}) that separation is not a suitable fairness requirement for risk score models.
    In addition, due to the differing risk distributions, one group or the other is highly over- or underrepresented in the selected set at different decision thresholds, relative to their respective target representation~$p(G \mid Y=1)$ (right panel).
    This illustrates how even a fair risk score model can lead to a potentially unfair ranking (see \cref{sec:ranking}).}
    \label{fig:rankingexample}
\end{figure}

Following the publication of various incompatibility theorems~\cite{Kleinberg2017, Chouldechova2017}, a misconception can sometimes be observed that calibration by group and error rate parity (also called equalized odds) cannot be achieved simultaneously, except in degenerate cases.
This is not true, however, and results from a misinterpretation of the original theorems: it is indeed possible to achieve both properties simultaneously, also in nontrivial cases~\cite{Reich2020, Hedden2021}.\footnote{The misconception results from applying one criterion -- calibration by groups -- to the risk scores~$R\in [0, 1]$ while applying the second criterion -- separation -- to binary predictions $\hat{Y}\in\{0, 1\}$ derived from the risk scores. The incompatibility theorem~\cite{barocas-hardt-narayanan} only holds when applying both properties to the same entity, i.e., either $R$ or $\hat{Y}$. For a longer discussion of this issue, refer to \citet{Reich2020}.}
What \emph{is} impossible to achieve is a risk model that simultaneously satisfies calibration by group and \emph{separation}, defined as $R \perp G \mid Y$~\cite{barocas-hardt-narayanan}.
Separation would require the conditional risk distributions~$p(R\mid G, Y)$ to be identical across groups, which is impossible to achieve if~$R$ is calibrated by groups and there are differences in the base rates~$p(Y=1 \mid G)$~\cite{Kleinberg2017}.
We mention this here, because an intuitively appealing way to achieve similar discriminative ability across groups might be to require identical risk score distributions~$p(R \mid G, Y)$ across groups, i.e., separation in terms of the risk scores~$R$.
In fact, the use of this fairness definition for risk scores has been proposed in the literature~\cite{Pfohl2019}.
As shown, however, this definition is in conflict with the aim of calibration by groups, which we consider central to risk score modeling.
It also penalizes predictors that fully conform with our aims: for example, a risk model can achieve equal AUROC scores in two groups and be perfectly calibrated by groups while violating separation, see \cref{fig:rankingexample} for an example.
To conclude, we emphasize that we do \emph{not} consider separation a suitable fairness criterion for risk scores, and that calibration by groups and equal discriminative ability by groups are fully compatible.

\vspace{0.8em}
\noindent Readers familiar with the standard algorithmic fairness literature may have noticed that our two desiderata outlined above -- calibration by groups and similar discriminative ability across groups -- closely match two of the three standard definitions of classification fairness: sufficiency, separation, and independence~\cite{barocas-hardt-narayanan}.
Calibration by groups implies sufficiency~\cite{barocas-hardt-narayanan}, and while we explicitly do not require separation (for reasons discussed above), that fairness conception could also be seen as an attempt to achieve similar discriminative ability across groups.
However, we did not discuss anything similar to \emph{independence} or \emph{demographic parity} so far.
This will change in the following section.

\section{Risk score-based fair ranking and resource allocation}
\label{sec:ranking}
In scarce resource distribution settings, such as medical triage, it is a natural idea to use risk models for informing decisions about resource distribution~\cite{Patel2009, Kutz2016}.
However, as we will outline here, such a procedure can lead to highly unfair distribution processes, even if the underlying risk score model is epistemically fair in the sense defined above.

For fair resource distribution, we will typically want to impose some requirement on the \emph{representation} of groups in the selected group, i.e., $p(G \mid \hat{Y}=1)$. 
As one example, we might require
\begin{equation}
    p(G \mid \hat{Y}=1) = p(G \mid Y=1) \quad \forall \,\, G,
    \label{eq:repr}
\end{equation}
that is, that the fraction of the selected individuals that is from group $G$ matches the fraction of individuals with outcome $Y=1$ that is from group $G$.
This is similar to the classical \emph{independence} requirement $R \perp G$, in the sense that we impose a requirement on the representation of groups in the selected set, regardless of the actual outcome~$Y$ of the selected individuals.
Since the amount of available resources is often unknown a priori or may change over time, we would want this to hold \emph{for any decision threshold~$\tau$}, at least in a certain threshold range $\Tau=[\tau_{\text{min}}, \tau_{\text{max}}]$ of interest.\footnote{This setting is similar to the one discussed by \citet{Vogel2021} and \citet{Zehlike2022b}. However, \citet{Vogel2021} consider error rate-based metrics, and \citet{Zehlike2022b} consider a setting in which relative ranking within the selected set matters.}

Unfortunately, the requirement outlined above is, in general, incompatible with the fundamental aim of risk modeling, i.e., to approximate the true risk~$\truerisk$ of individuals as accurately as possible.
Since the expression on the right-hand side of \cref{eq:repr} is threshold-independent, this would require that the risk score distributions $p(R \mid G)$ be proportional to one another within the threshold range~$\Tau$.
However, the distribution of true risks~$p(\truerisk \mid G)$ may differ between groups, and in particular, they may not be proportional to one another within the range~$\truerisk \in \Tau$.
Thus, the outlined representation requirement is in conflict with the fundamental aim of risk modeling: we can either optimize for risk prediction accuracy, or for fair ranking, but doing one will detract from the other.
A simple illustration of this issue and its consequences for (un)fair resource access prioritization is shown in~\cref{fig:rankingexample}.

Of course, a risk score can still be used as the \emph{basis} for a fair ranking scheme that fulfils \cref{eq:repr} for all thresholds of interest, using various methods described in the score-based fair ranking literature~\cite{Vogel2021, Zehlike2022, Zehlike2022b}.
However, the result of such a process would no longer represent a risk model itself, since its scores would have been adjusted in a manner that optimizes representation fairness while possibly detracting from risk accuracy.
This means that while a fair ranking can be \emph{derived} from a fair risk score, ranking \emph{directly} based on the predicted risk scores is likely to be unfair.

\section{Assessing the fairness of risk score models}
\label{sec:metrics}
We now turn our attention to the empirical quantification of risk model fairness.
In \cref{ssec:groups}, we lay out fundamental requirements on metrics to be used for such assessments, most of which derive from the observation that in fairness analyses, groups of different \emph{sizes} are compared.
Based on these requirements, we then lay out our choice of specific metrics in \cref{ssec:metrics}.
The code for all our metric implementations is available online.\footnote{A link to the code will be added here after acceptance.}

\subsection{Comparing metrics between groups}
\label{ssec:groups}
In almost all practical fairness analyses, groups of different \emph{sizes} will be compared, with important methodological consequences.
First, metrics computed on small test groups will generally suffer from higher uncertainty compared to larger groups.
In order to perform a proper comparison between groups, \emph{quantifying} this uncertainty is crucial.
As we observe in our case studies in \cref{sec:case-studies}, metric uncertainty in small groups is often so high that statements concerning unfairness cannot be made with any certainty.
Second, and at least as importantly, comparing groups of different sizes requires \emph{sample size-unbiased} metrics.
For instance, standard calibration error metrics exhibit a strong sample size bias, returning higher values in smaller groups even when predictions are equally well-calibrated~\cite{Kumar2019, Roelofs2022, Gruber2022}.
This is clearly problematic when using such a metric for comparing model performance between groups of different sizes -- as is widely done -- since one might wrongly conclude that groups are differently well-calibrated when the underlying calibration quality is the same but group sizes differ.
We will investigate this sample size bias and alternative, less biased ways of quantifying calibration errors in \cref{ssec:calibration}.
An additional requirement on metrics to be used in fairness analyses results from different groups having different \emph{base rates}~$p(Y \mid G)$:
in order to meaningfully compare metrics between groups, metrics must be -- at least to a large degree -- invariant with respect to varying base rates.

Finally, how to perform the actual \emph{comparisons}?
Researchers have proposed, for example, pair-wise group comparisons, comparisons to the best-performing group in each metric, and comparisons to the complement of each group, i.e., all individuals that are \emph{not} members of the group under consideration.
Comparisons have been performed by assessing, for example, metric differences or metric ratios between groups.
In the interest of obtaining supposedly objective statements regarding the presence or absence of unfairness, one might also consider statistical hypothesis testing regarding group differences.
However, the challenges surrounding the use of statistical hypothesis testing and p-values are well-documented~\cite{Greenland2016}.
Ultimately, the domain expert will need to decide which level of differences between groups, and at which level of certainty, is considered problematic.
In order to maximize information flow to the domain expert, we thus argue for simply reporting confidence intervals for each metric and group separately, instead of any derived (e.g., difference or ratio-based) quantities.
For the same reason, we consider it essential to report not just a single metric value, but to accompany each metric with a detailed graphical assessment that enables a more detailed assessment.

\subsection{Choice of specific metrics}
\label{ssec:metrics}
Throughout the previous sections, we have outlined a number of requirements for metrics to be used in fairness comparisons.
These included, in particular,
\begin{enumerate}
    \item orthogonality (calibration metrics should be independent of discriminative ability, and vice versa),
    \item sample size-unbiasedness (the expected value of a metric should not depend on the size of the test sample), and
    \item independence of group base rates (it should be meaningful to compare groups with different base rates).
\end{enumerate}
Finally, to reiterate, we are interested in metrics that assess risk scores~$R\in[0, 1]$, not binary classifications~$\hat{Y} \in \{0, 1\}$.

\subsubsection{Calibration performance metrics}
\label{ssec:calibration}
Following our discussion of calibration error metrics and their sample size biases in \cref{sec:background}, we here propose a combination of two previously described methods, which we show to further reduce sample size bias and metric variability compared to both of them.
In particular, we propose to combine the debiasing method of \citet{Kumar2019} with an adaptive bin count search method similar to the one described by \citet{Roelofs2022}.
To find the largest bin count that still yields a monotonic calibration curve, \citet{Roelofs2022} iteratively increase the number of bins~$N_B$ one by one until the calibration curve becomes non-monotonic.
Here, to speed up convergence, we use a simple interval search method for~$N_B$ instead.
Moreover, we ensure that each bin always contains at least~10 samples.
We accompany our calibration analyses with loess-based reliability diagrams~\cite{Austin2013} using bootstrap-derived confidence intervals~\cite{Austin2014}.
Further metrics could be considered, such as the calibration intercept for assessing whether risks are on average over- or underestimated in a group~\cite{Calster2019}.

\begin{figure}[t]
    \includegraphics[]{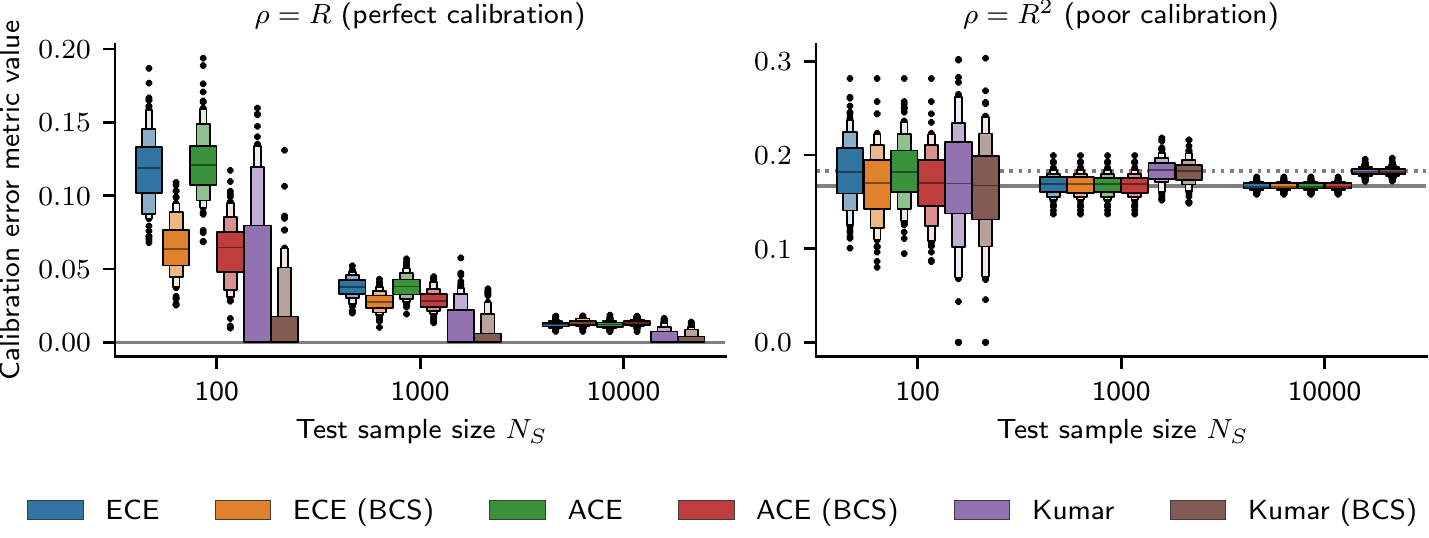}
    \caption{Analysis of the bias of different calibration error metrics with respect to test sample size~$N_S$.
    Model predictions~$R$ are drawn from a standard uniform distribution; true risks are then given by $\rho=R$ (perfectly calibrated) or $\rho=R^2$ (poorly calibrated), and binary outcomes~$Y$ are drawn from a Bernoulli distribution with risk $\rho$.
    Horizontal lines indicate the ground-truth expected calibration error (solid) and the ground-truth root mean squared calibration error (dotted).
    Each experiment is repeated 100 times, and a boxenplot~\cite{Hofmann2017} over the distribution of the resulting metric values is shown.
    For a clarification of the compared metrics, refer to the main text (\cref{sec:background} and \cref{ssec:calibration}); ``(BCS)'' denotes our adaptive bin count search method.
    Note the different scaling of the y axes.
    }
    \label{fig:samplesizebias}
\end{figure}

\Cref{fig:samplesizebias} shows the distribution of calibration metric values in two simple synthetic test cases, one perfectly calibrated and one poorly calibrated.
In agreement with previous publications~\cite{Roelofs2022, Gruber2022} we find that standard ECE and ACE (with a fixed number of~$N_b=15$ bins) exhibit strong biases especially for well-calibrated models and small sample sizes, and that both adaptive bin count search and calibration error debiasing as proposed by \citet{Kumar2019} improve metric reliability.
In particular, the combination of the debiased root mean square calibration error (DRMSCE) proposed by \citet{Kumar2019} with our adaptive bin count search described above outperforms all other metrics.
We will simply call this combination DRMSCE and use it in the two case studies presented in \cref{sec:case-studies}.

\subsubsection{Discriminative ability}
\label{ssec:discrimination}
We use AUROC as our main metric to quantify discriminative ability, as it is independent of calibration performance, sample size-unbiased, and largely independent of group base rates.
In case of strong class imbalance, i.e.,~$p(Y) \ll 0.5$, we perform precision-recall analysis instead, as recommended in the literature~\cite{Saito2015, MaierHein2022}.
In order to enable meaningful comparisons between groups of different base rates, we use the area under the precision-recall-\emph{gain} curve (AUPRG) in this case, which represents a properly normalized AUPR~\cite{Flach2015}.
We accompany our analyses of discriminative ability with ROC and PRG curves, including bootstrap-based confidence intervals.

\subsubsection{Ranking fairness}
In order to quantify ranking fairness according to the criterion proposed in \cref{sec:ranking}, we introduce the \emph{expected under-representation}, which we define as
\begin{equation}
    \mathrm{EUR}(G) = \mathrm{E}_{\tau \sim p_{\text{emp}}(R)} \left\{ \min \left( \frac{p(G \mid \hat{Y}(R; \tau) = 1)}{p(G \mid Y=1)}, 1\right)\right\},
    \label{eq:eur}
\end{equation}
where~$\hat{Y}(R; \tau)$ denotes binary decisions derived from~$R$ using threshold~$\tau$, and $p_{\text{emp}}(R)$ denotes the empirical distribution of risk scores in the test set across all groups.
The ranking fairness analyses are accompanied by a plot of the normalized group representation
\begin{equation*}
    \frac{p(G \mid \hat{Y}(R; \tau) = 1)}{p(G \mid Y=1)}
\end{equation*}
as a function of the decision threshold~$\tau$, again using bootstrap-based uncertainty quantification.

\section{Case studies}
\label{sec:case-studies}
We present two real-world case studies to illustrate our fairness assessment methodology.

\subsection{Experimental methodology}
Datasets are split into training, validation, and test sets, and we train a standard XGBoost~\cite{Chen2016} classification model on the training set.
We perform randomized hyperparameter search using five-fold cross-validation on the training set and calibrate the resulting model using beta calibration~\cite{Kull2017a} (first case study, because of relatively small validation set size) and isotonic regression~\cite{Zadrozny2002} (second case study), respectively, on the validation set.
We select a set of sensitive attributes to consider (e.g., gender, age group, etc.), and perform our further analyses for all groups that can be constructed from value combinations of these attributes (say, elderly women), and that surpass a minimum group size threshold. 

Performance is evaluated on the hold-out test set, and we use a simple test set bootstrapping method to quantify metric uncertainty.
From the overall test set, we select the~$N_g$ samples belonging to group~$g$.
We then draw~$N_B=200$ bootstrapped test sets (each of size $N_g$) from this original group-specific test set, compute the metric of interest (e.g., AUROC or calibration error) on each of them, and use the obtained distribution of metric values for quantifying the uncertainty associated with this metric on this particular group.
In particular, we will consider the median and the (percentile-based) 95\% confidence intervals for further analysis.
The code for our analyses is available online.\footnote{The link to the code will be added here after peer review.}

\subsection{Case study A: Catalan juvenile recidivism risk assessment}
\label{ssec:catalan}
We use a dataset on juvenile recidivism provided by the Centre for Legal Studies and Specialised Training (CEJFE) within the Department of Justice of the Government of Catalonia,\footnote{The dataset can be downloaded at \url{https://cejfe.gencat.cat/en/recerca/opendata/jjuvenil/reincidencia-justicia-menors/index.html}.} analyses of which have been previously published by \citet{Tolan2019} and \citet{Fuglsang2022}.
The dataset provides information about juvenile subjects who have participated in an educational program following a conviction of a criminal act in Catalonia.
We predict the risk of recidivism within five years after completion of the educational program, based on subject age at time of the crime and at time of program completion, sex, area of origin, province of residence, the number of prior criminal records, the province in which the educational program was executed, crime category and specific type, and the type of educational program.
We use the preprocessing described by \citet{Fuglsang2022}, available online.\footnote{\url{https://github.com/elisabethzinck/Fairness-oriented-interpretability-of-predictive-algorithms/blob/main/src/data/cleaning-catalan-juvenile-recidivism-data.py}}
The processed dataset contains 4652 samples (34\% recidivism rate), and we use 70\% / 10\% / 20\% for the training / validation / test sets, stratifying on the recidivism outcome.
We consider subject sex, age group at time of crime, subject area of origin, as well as the province of residence as sensitive variables for our fairness analyses, and we use a minimum test set group size of 100 for our fairness analyses, leaving a total of 41 groups to be analyzed.

\begin{figure}[p]
    \includegraphics[width=\textwidth]{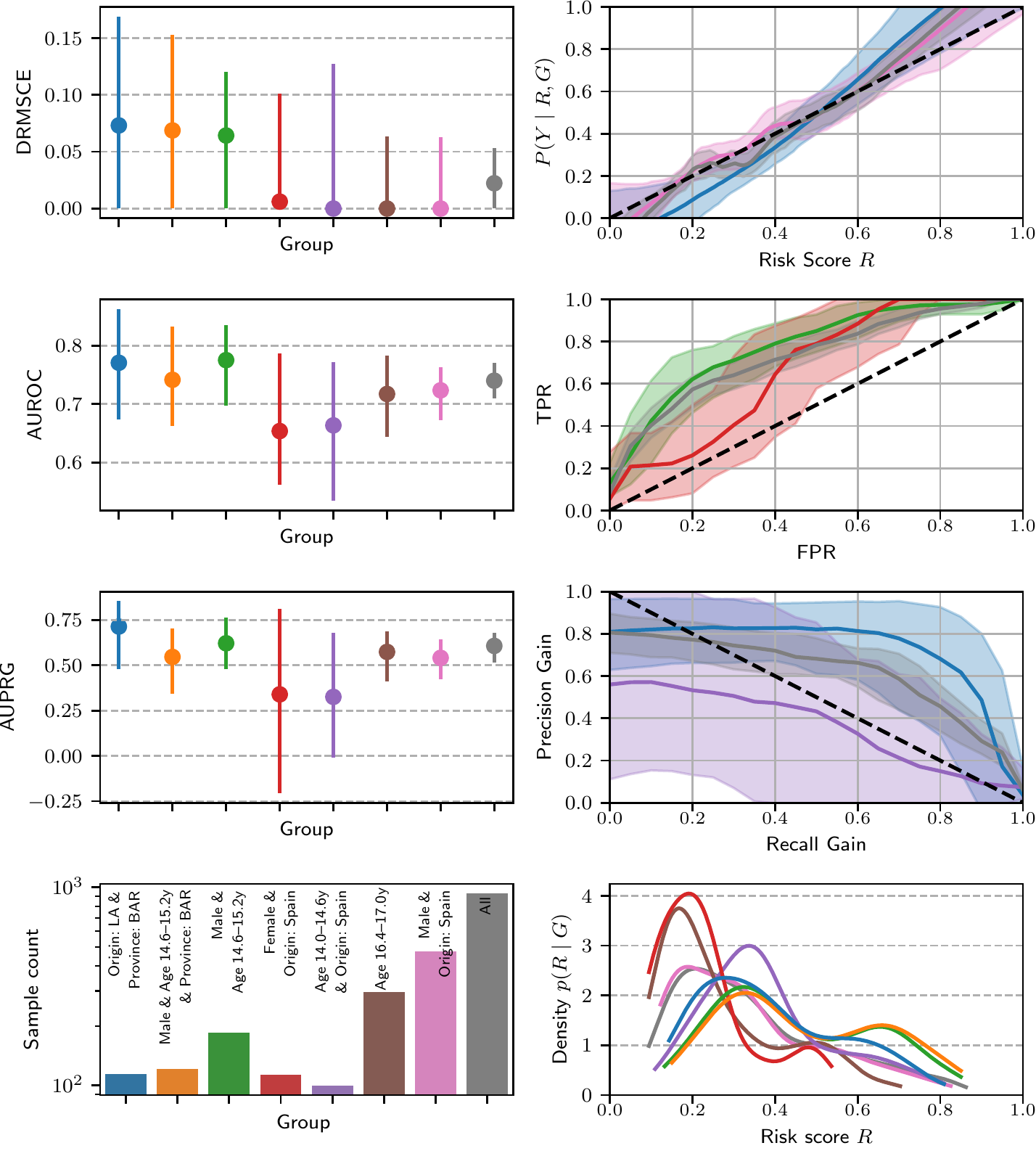}
    \caption{Analysis results for the Catalan juvenile recidivism dataset, on a few selected subgroups.
    The left column shows the distribution of metric values obtained by test set bootstrapping as described in~\cref{ssec:groups}.
    For details regarding the different metrics, refer to \cref{ssec:metrics}.
    The right column shows (top to bottom) a loess-based reliability diagram~\cite{Austin2013, Austin2014} (with the best and worst calibrated groups according to DRMSCE), a receiver-operating characteristic (ROC) diagram (with the best and worst-performing groups according to AUROC), a precision-recall-gain diagram (with the best and worst-performing groups according to AUPRG), and the distribution of risk values in the different groups.
    The same colors are used for all groups in all panels, cf. the bottom left plot for a key.
    DRMSCE is equivalent to Kumar (BCS) in \cref{fig:samplesizebias}, LA=Latin America, and BAR=Barcelona.
    Age refers to age at time of crime.
    }
    \label{fig:catalan}
\end{figure}

\Cref{fig:catalan} shows the results of our analyses.
Overall model performance is reasonable (AUROC~>~0.7, DRMSCE~<~0.05) given the limited amount of input information.
As expected, metric uncertainty in small test groups is very high, making reliable statements about the fairness or unfairness of the model near-impossible.

\subsection{Case study B: National registry depression risk assessment}
Major Depressive Disorder (MDD) is considered the most burdensome disease in Europe~\cite{Wittchen2011} and the leading cause of disability globally~\cite{WHO2017}.
To address this challenge, depression risk assessments have been proposed for early detection and prevention~\cite{Voorhees2008, Moriarty2020, Fried2022a}.
At the same time, public authorities and health insurance companies are actively exploring ML-based healthcare resource prioritization, in mental health and other domains~\cite{Obermeyer2019, Pnevmatikakis2021, Joyce2021}.
In this case study, we aim to investigate potential algorithmic fairness challenges that arise when using a depression risk assessment to decide who gets access to limited healthcare resources, and who does not.

We use data from a national registry database\footnote{To maintain full anonymity during peer review, we have removed any information about the country from which data were collected.} that includes demographic and healthcare status information of the national population aged 15 years or older per the 31st of December from 2000--2018.
For each year, we include all subjects diagnosed by a medical practitioner with MDD (ICD codes DF32 and DF33) within that year as cases.
Control subjects were sampled without replacement from those who were not diagnosed with MDD that year, were alive at the end of the year, and were not previously selected as cases.
Two equally sized cohorts without subject overlap were constructed this way, one for training, and one for validation and testing, both balanced between cases and controls and containing approximately 240,000 subjects each.
The latter set was split into validation (1/3) and test (2/3) sets.
We train a model to predict MDD diagnosis from demographic information; a complete list of predictors and sensitive variables can be found in \cref{sec:dst-details}.
We analyze model performance in groups with a test set size of at least 2,000 subjects.

\begin{figure}[p]
    \includegraphics[width=\textwidth]{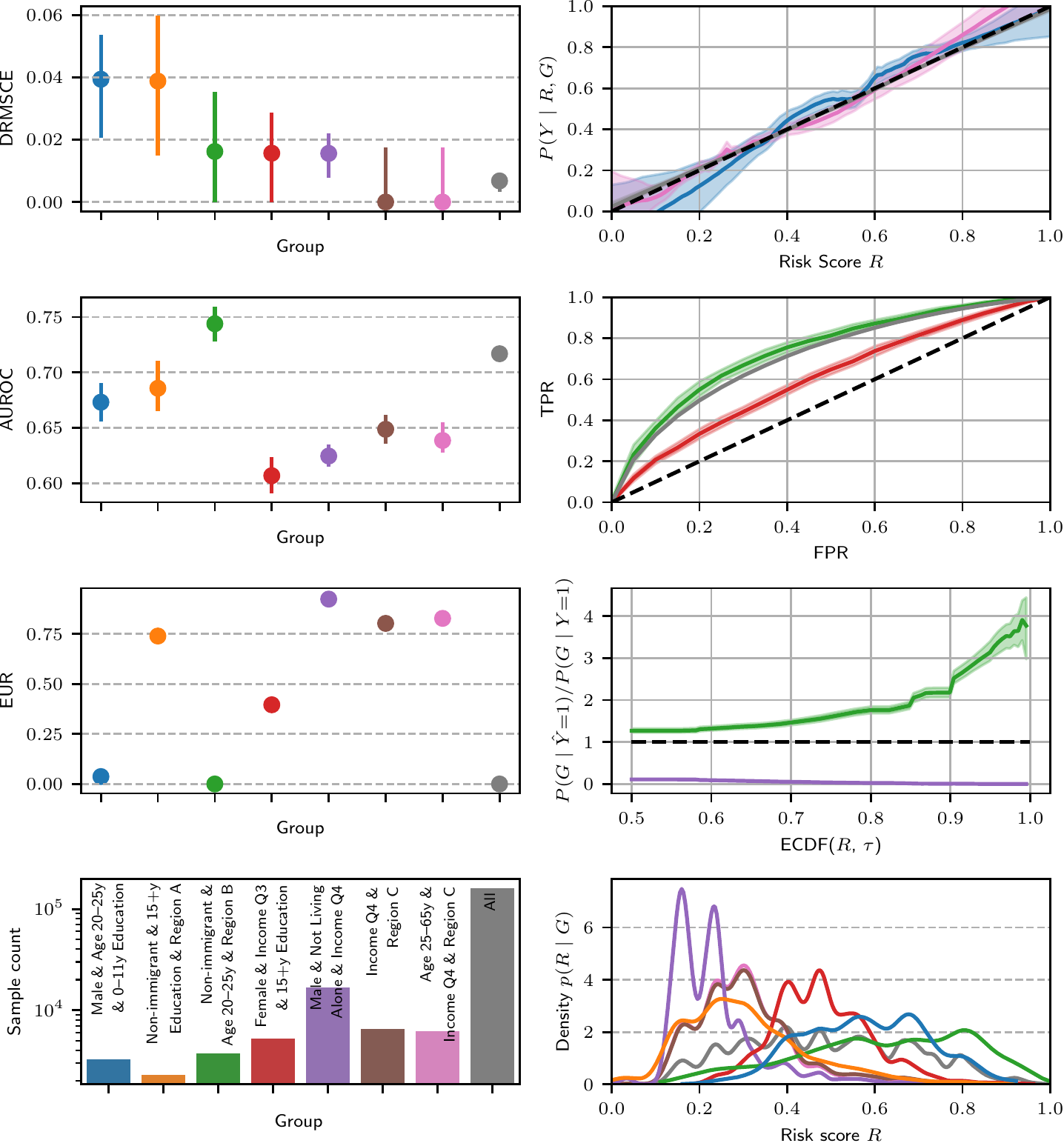}
    \caption{Analysis results for the depression dataset, on a few selected subgroups.
    The left column shows the distribution of metric values obtained by test set bootstrapping as described in~\cref{ssec:groups}.
    For details regarding the different metrics, refer to \cref{ssec:metrics}.
    The right column shows (top to bottom) a loess-based reliability diagram~\cite{Austin2013, Austin2014} (with the best and worst calibrated groups according to DRMSCE), a receiver-operating characteristic (ROC) diagram (with the best and worst-performing groups according to AUROC), group representation~$p(G \mid \hat{Y}{=}1)$ as a function of the decision threshold~$\tau$, and the distribution of risk values in the different groups.
    The same colors are used for all groups in all panels, cf. the bottom left plot for a key. 
    Income QX denotes equivalized household income quartile X, cf. \cref{sec:dst-details} for details.}
    \label{fig:dst}
\end{figure}

\Cref{fig:dst} shows the results of our analyses.
Overall model performance is again reasonable (AUROC~>~0.7) and especially overall calibration is very good (DRMSCE~<~0.01).
Moreover, since this dataset is much larger, more reliable statements regarding potential disparities can be made.
Some groups are clearly better calibrated than others; the groups that are poorer calibrated tend to be small.
There are also large differences in discriminative ability, with AUROCs ranging from 0.6 to 0.75 in different groups.
(We did not consider AUPRG here since the dataset is perfectly outcome-balanced.)
This is also reflected in measures of under-representation in the selected set: subjects from groups with higher AUROCs tend to receive more extreme risk estimates, resulting in them being selected earlier compared to subjects in groups with lower AUROCs.

\section{Discussion, limitations, and open challenges}
In the following, we will discuss important outcomes and limitations of our analyses and case studies.

\subsubsection*{Challenges posed by varying test sample sizes.}
While the issue of varying \emph{training} set sizes for different groups has received widespread attention in the fairness literature, challenges posed by varying \emph{test} set sizes are less widely discussed.
In small test groups, such as in our juvenile recidivism case study, metric uncertainty is often so high that statements about fairness or unfairness cannot be made with any degree of certainty -- and this is even \emph{after} excluding groups below a minimum group size.
It should thus be emphasized that ensuring broad group representation in the test set is at least as important as ensuring representation in the training set.
Concerning metric evaluation, it is paramount that researchers take sample size variability into account in their analyses and use proper metric uncertainty estimates as well as sample size-unbiased metrics, such as the new calibration error metric we propose here.

\subsubsection*{Choice of metrics}
We do not aim to make specific metric recommendations here; rather, our aim was to elucidate the basic requirements that any appropriate metric will need to fulfill (see \cref{ssec:metrics}).
The choice of the most appropriate metrics to consider will always be application-dependent~\cite{MaierHein2022}.
For instance, AUROC and AUPRG take the whole threshold range into account -- in many applications, however, a smaller range of thresholds may be of interest, thus rendering partial AUCs more appropriate~\cite{McClish1989}.
Various other variations on ROC-based and ranking metrics have also been proposed~\cite{Vogel2021}, such as xAUC~\cite{Kallus2019}, which measures the likelihood that a random positive example from one group will be ranked above a random negative example from another group.

\subsubsection*{Performance differences and unfairness mitigation.}
In our case studies, we observe that calibration and discriminative performance are, indeed, independent properties: groups can perform well on both, either of them, or none of them.
At the same time, we observe, like many others before us, that our models perform better on some groups compared to others.
What are the origins of such performance differences?
Saving a more complete discussion of this question for elsewhere, we briefly remark that performance differences can result from a number of root causes, including group under-representation~\cite{Larrazabal2020}, algorithmic choices~\cite{Hooker2021, Forde2021}, missing covariate information that is more important for some groups compared to others, and varying levels of input or label noise across groups~\cite{Zhang2022}.
While some of the resulting performance differences can be reduced by means of algorithmic interventions, this has generally proven to be a challenging endeavour.
A number of studies have found that algorithmic solutions to fairness problems often yield undesirable or only marginally improved results~\cite{Pfohl2021, Pfohl2022}.
This is not surprising: if the underlying problem is that there are simply too few samples from a group, or that data from one group are more noisy than those from others, purely algorithmic interventions seem unlikely to solve the issue.
Instead, an investment in more and better data from currently under-performing groups will often be the more promising approach.

\subsubsection*{Risk scores and fair ranking.}
We have discussed the relationship between fair risk scores and fair ranking, finding that while a fair ranking can be \emph{derived} from a fair risk score, ranking \emph{directly} based on the predicted risk is likely to be unfair (see \cref{fig:rankingexample}).
Another interaction between discriminative ability and ranking can be observed in our second case study (see \cref{fig:dst}).
Groups in which prediction is harder (as indicated, for instance, by lower AUROC) tend to receive less extreme risk estimates and are, thus, likely to be under-represented in a risk-based set selection scheme.
Following our findings, one might consider it more appropriate to report both an estimated risk score \emph{and} a suggested ranking based on the computed risks.
This, of course, raises further challenges of real-world human--model interaction, many of which are yet to be addressed even for the ``simple'' risk-only case~\cite{Green2019, Rambachan2020, Green2021, Simon2021, Yarborough2022, Mantell2021}.
In addition, there are further complexities involved in fair ranking for resource prioritization, as we will discuss next.

\subsubsection*{Artificially quantized quantities.} 
Our study -- like most of the current literature -- is based on the assumption of discrete outcomes.
However, real-world outcomes often do not conform with discrete categories (``disease present'' -- ''disease absent'').
Instead, disease severity is often a \emph{continuous} property that is artificially discretized during the data collection procedure.
Such artificial discretization of continuous variables is well-known to confound statistical analyses~\cite{Royston2005, Altman2006, OakdenRayner2020a}, and it might also cause problems in our setting.
For example, while disease \emph{incidence} might be similar in two groups, disease \emph{severity} could be higher in one group compared to the other, with important consequences for what would be considered a ``fair'' ranking.
Such differences are brushed over by approaches relying on discretized outcomes.

\subsubsection*{Biased data.}
One crucial limitation of our study and most other approaches to algorithmic fairness is its underlying assumption that the available data are unbiased~\cite{Friedler2021}.
All of our metrics -- like most metrics currently used in the literature -- can hide arbitrary degrees of unfairness if the dataset suffers from, for instance, selection biases~\cite{Kallus2018, Bao2022}, label biases~\cite{Zhang2022, Bao2022}, or the choice of a biased proxy variable~\cite{Obermeyer2019}.
Further aggravating the problem, such biases are often undetectable without deep domain knowledge or further data collection efforts.
Like others~\cite{Jacobs2021}, We consider this one of the fundamental challenges of algorithmic fairness, and we highly appreciate attempts~\cite{Obermeyer2019, Madras2019, Fogliato2020, Wang2021a} to address this problem.
Even leaving objective measurement biases aside, subjective \emph{choices} are always involved in the decision to measure a quantity, label something a crime or a disease, or include certain subjects but not others~\cite{McCradden2020, Biddle2020, Bao2022}.
In this sense, it is crucial to realize that risk models can never be fully \emph{neutral} or \emph{objective}.

\subsubsection*{Fairness analyses in context.}
We here judge the fairness of a risk model purely based on the joint distribution of generated risk scores~$R$, group membership~$G$, and output labels~$Y$, ignoring questions of interpretability~\cite{Liu2022, Loreaux2022, Crabbe2020}, procedural fairness~\cite{Lee2019}, algorithmic recourse~\cite{Karimi2021}, or wider societal context~\cite{McCradden2020, Biddle2020}.
It is known, for example, that some demographic groups suffer from worse access to healthcare services compared to others. 
``Fairly'' equalizing the access of groups to a particular healthcare service, while disregarding such discrepancies in baseline healthcare access, might contribute to deepening existing healthcare disparities.
Decisions on whether or how to deploy an algorithm in the real world should thus never be based purely on a narrow fairness analysis such as ours.
Instead, a comprehensive assessment of real-world complexities is essential, in which an analysis using our methodology can play just a small part~\cite{Selbst2019, Green2019, Rambachan2020, Green2021, Zicari2021, Moss2021}.
Nevertheless, fine-grained analyses of the different places in which unfairness can occur are crucial to navigate this immensely complex decision space, and we hope our work to be valuable towards this end.

\section{Conclusion}
Our main argument in this paper has been that the fairness of risk score models should be evaluated \emph{on their own}, that is, independent of any subsequent decision-making processes.
The reasons for this perspective were twofold.
First, separating between the fairness of a \emph{model} and the fairness of \emph{decisions} made based on that model enables a more fine-grained analysis and discussion about the sources of unfairness.
And second, risk models are indeed often used to provide information to human decision-makers, or at time-varying decision thresholds, in which case it is even more crucial to assess the model's fairness on its own.
We have further argued that the fairness of risk score models then is a function of the \emph{epistemic value} they provide to different groups, and we have proposed metrics to empirically quantify this epistemic value, including a novel, debiased calibration error metric.
Finally, we have considered the fairness of risk-based resource prioritization, finding that a ranking of subjects based purely on predicted risks will often be unfair.

\subsubsection*{Acknowledgements}
The authors would like to thank Vibe Frøkjær, Neurobiology Research Unit, Copenhagen University Hospital, Rigshospitalet, and Merete Osler, Department of Public Health, University of Copenhagen, for help with the MDD study conception.
The authors would also like to thank Emily Beaman, Department of Biomedical Sciences, University of Copenhagen, for assistance with the data analysis in the MDD study.

\subsubsection*{Funding sources}
Work on this project was funded by the Independent Research Fund Denmark (DFF), grant number 9131-00097B.
The funding agency had no influence on the writing of this manuscript, nor on the decision to submit it for publication.
All authors are affiliated with the Pioneer Centre for AI, DNRF grant number~P1.

\FloatBarrier

\printbibliography

\newpage

\appendix

\section{Additional details on the MDD case study}
\label{sec:dst-details}
Subjects may be selected as a control in one year and as a case in a later year (in which they were diagnosed with MDD), although this occurs only rarely in the dataset.

The used predictor variables were the following:
\begin{itemize}
    \item age, 
    \item gender (male / female / changed at some time in the available dataset)
    \item registered identification as LGBT, as detected by either a registered same-sex partnership or sex change,
    \item living status (living alone or not),
    \item equivalized household income quartile (accounts for number and age of children and combined household earning), 
    \item years of school and university education, 
    \item heritage (immigrant, child of immigrant parents, non-immigrant),
    \item civil status (married, not married, divorced, widowed),
    \item number of address changes during the year,
    \item larger region of residence,
    \item smaller region of residence.
\end{itemize}
As sensitive variables for our group fairness analyses, we consider age group, gender, registered identification as LGBT, heritage, larger region of residence, equivalized household income quartile, living status (alone or not), and number of years of education.
We consider groups resulting from the combination of at most three sensitive variables and with a test set size of at least 2,000 subjects, resulting in the analysis of model performance in 886 different groups.
\Cref{fig:dst} in the main paper shows results for eight of those.

\end{document}